# Zero-shot evaluation of open vision-language models as System 1 decision models on radiology tasks

Bradley J Erickson & Sara Salehi

Mayo Clinic AI Laboratory

Rochester, MN 55902 USA

{ bje, salehi.sara } @mayo.edu

## Abstract

**Purpose.** To compare two multimodal System 1 decision models, Clef 27B and d1 3B, with three generative vision-language models read out zero-shot in single-token output mode (Clef's backbone Qwen3.8-27B, Gemma 4 31B and the medically tuned MedGemma 1.5 4B) on seven radiology tasks covering text, radiographs, CT and MRI.

**Methods.** Five models were run locally on a single NVIDIA DGX Spark (GB10): Clef, a 27B System 1 decision model based on Qwen3.8-27B; its backbone Qwen3.8-27B; d1, a 3B System 1 decision model; Gemma 4 31B; and MedGemma 1.5 4B, a model tuned on medical text and images. Clef and d1 returned a probability for every option in one non-autoregressive pass. The three generative models were made to answer with a single option letter, and the probability distribution over the option letters specified in the prompt was taken as their output. Three tasks were interpretation of text and 4 were detection of finding in images. The tasks were: 1. finding presence of finding in CT-RATE chest CT reports (n = 1,000, 18 findings); 2. Presence of finding in PadChest chest radiograph reports (n = 1,000, 20 findings); 3. the same 18 findings on CT-RATE chest CT but 3D volumes (n = 100); 4. MR safety status of devices from FDA AccessGUDID device records (n = 999); 5. line and tube position on RANZCR CLiP chest radiographs (n = 500, 11 labels); 6. motion-artifact severity on MR-ART T1-weighted brain MRI (n = 436) and 7. sequence identification on BraTS 2021 brain MRI (n = 400). The primary endpoint was AUROC with a 95% case-bootstrap confidence interval. These are all 'Zero-shot'--there was no fine-tuning of the models on the tasks.

**Results.** All five models completed all 4,435 tests without a failed request. All five retrieved findings from reports well (AUROC 0.93–0.99). Clef ranked higher than its Qwen backbone on CT-RATE volumes (0.749 vs 0.668; paired Δ +0.082, 95% CI +0.045 to +0.119) and RANZCR (0.812 vs 0.766; Δ +0.046, +0.021 to +0.074) and lower on MR-ART (0.625 vs 0.709; Δ −0.084, −0.103 to −0.065). The two did not differ on BraTS, GUDID or PadChest, and differed by 0.002 on CT-RATE reports. Clef changed decisions more than ranking. On BraTS it correctly identified 90 of 100 FLAIR volumes, a class its backbone never chose, and accuracy rose from 0.615 to 0.888. On CT-RATE volumes and RANZCR, its backbone answered "yes" to 88% and 65% of finding questions against prevalence of 17% and 13%. Clef answered "yes" to 19% and 3%, and expected calibration error fell from 0.436 to 0.072 and from 0.255 to 0.025. The 3B model d1 ranked lower than Clef on six of seven tasks, by 0.03–0.04 on reports and by 0.13–0.28 on BraTS, GUDID, CT-RATE volumes and RANZCR. Clef, its backbone and d1 made no usable decisions on motion severity: each labelled at least 99% of MR-ART scans "none or minimal". Clef was 1.5 to 2.9 times faster per request than its backbone on tasks with 18 or 20 questions per input, where one pass replaced many calls, but slower on single-question tasks. d1 used 3 to 9 times less energy per request than Clef. Of the two added generative models, Gemma 4 31B ranked MR safety best of all five models (AUROC 0.854 vs 0.762 for Qwen3.8-27B) but was lower than Qwen3.8-27B on report text (0.933–0.942), and MedGemma 1.5 4B ranked motion severity best (0.821) but answered "severe" for 93% of scans and was at chance on MRI sequences (0.495).

**Conclusion.** System 1 models can perform important tasks in radiology with high accuracy, at far lower compute cost than LLMs used in the usual way, with reasoning and free-text answers. When an LLM is constrained to a single answer token, its cost per question is similar to System 1 models, and a System 1 model is faster only when it answers several questions in one pass. A System 1 head on the same backbone added little discrimination on report text, but it gave more usable decisions on MRI sequences and better calibration on chest radiograph and CT findings. A 3B System 1 model was close to the 27B models on report text but not on images. No model made usable decisions on motion severity, and rare findings still needed per-label thresholds for satisfactory performance. The single-token readout transferred without changes to Gemma 4 and MedGemma, but no generative model was best on every task, and each had its own option bias. Probability readouts, whether from a dedicated head or a general-purpose model, need local calibration before they can act as decision components in radiology.

## Introduction

Many radiology decisions are a choice among a few fixed options made in large numbers: which sequence a series is, whether a line is in the right place, whether motion makes a scan non-diagnostic, and whether an implant is safe for MRI scanning. A model that makes such decisions must also say how certain it is, so that confident answers can be acted on and uncertain ones routed to a radiologist. Generative language models answer in free text, often after ‘reasoning’, which is slow, costly and gives poorly calibrated confidence. System 1 models instead return a probability for every option in a single forward pass through a model.

Decision models, also known as ‘System 1’ models, have recently gained attention (https://docs.typesafe.ai/introduction) because of their speed and reduced cost for making decisions as opposed to generating free text. Yang, Vrettos et al. recently evaluated four System 1 models zero-shot on 86 public tasks across medicine (arXiv:2610.31315). They found that findings stated in radiology reports were retrieved almost perfectly (AUROC up to 1.00), while performance fell with each step from radiograph perception (mean AUROC 0.89) to 2D findings (0.84) to 3D CT and MRI volumes (0.69). Three of their tasks are repeated here: MR safety status of devices, their weakest text task (best AUROC 0.66), MRI motion-artifact severity (about 0.71), and line and tube position on chest radiographs. They also compared each open System 1 model with its own generative backbone. The backbones were 3 to 5 percentage points more accurate but worse calibrated, and 11 to 76 times slower.

In this manuscript, we refer to the “backbone**”** as the pretrained vision-language model, here Qwen3.8-27B. It turns the image and text into an internal representation. The “head” is the final stage that turns that representation into an answer, and the two models’ heads work differently:

- **Generative head:** predicts the next token over its whole vocabulary. Each token (~ word) generated requires at least one pass through the model (more with reasoning). To get a forced choice as a single token out, you prompt it to answer using one of the choices and read the probabilities it gives each option. That’s the readout used for Qwen3.8-27B, Gemma4-31B and MedGemma1.5-4B.
- **System 1 decision head (Clef, d1):** predicts among the question’s declared options directly, returning one probability per option in a single pass. It can also answer multiple questions in a single pass.

This manuscript addresses three questions open for radiology. First, does a dedicated decision head perform better than its backbone, when the backbone's thinking is turned off and it generates a single answer token? Second, can a much smaller (3 billion parameters) System 1 model stand in for a larger one (27 billion parameters)? Third, how much does the choice of backbone matter? The open System 1 model Clef is fine-tuned from Qwen3.8-27B, so the two can be run side by side on the same weights. This study compared Clef, Qwen3.8-27B and the 3B System 1 model d1, together with two further generative models read out the same way (Gemma 4 31B and the medically tuned MedGemma 1.5 4B), on seven tasks: finding presence in chest CT reports, chest radiograph reports and chest CT volumes, MR safety from device records, line and tube position on chest radiographs, motion severity on brain MRI, and sequence identification on brain MRI. The question was whether the head, model size or a newer backbone improves discrimination, and whether better discrimination carries over to usable decisions. We also describe the use of a traditional generative large language model (LLM) running in a single output token mode rather than the usual free text output form.

# Methods

## Tasks and datasets

Seven tasks were evaluated. The harness follows the zero-shot method of Yang, Vrettos et al. (*The Value of System 1 Models in Medicine*, arXiv:2610.31315). Each case is an input (text, images or both) with one or more typed questions, and the model returns a probability for every option. Cases were drawn with a fixed seed (42), so all models saw exactly the same cases and inputs. Code will be provided at: https://github.com/slowvak/System1Benchmark upon acceptance in a peer-reviewed publication.

| Task | Tier | Dataset | Cases | Question and options | Reference standard |
|---|---|---|---|---|---|
| CT-RATE reports | Retrieval from text | CT-RATE, validation split | 1,000 reports, 18 questions each (18,000 answers) | Yes/no per finding: is it described in the report? | Labels extracted from the report by the dataset's text classifier |
| PadChest reports | Retrieval from text | PadChest, physician-labelled reports (Spanish) | 1,000 reports, 20 questions each (20,000 answers) | Yes/no per finding (20 most frequent) | Physician labels of the report |
| GUDID MR safety | Judgement from text | FDA AccessGUDID, full delimited release | 999 (333 per class) | MR Safe / MR Conditional / MR Unsafe | Manufacturer-reported MRI safety status |
| RANZCR lines | 2D findings | RANZCR CLiP, training split | 500 radiographs, 11 questions each (5,500 answers) | Yes/no per line or tube label | Public competition labels |
| CT-RATE volumes | 3D findings (volume) | CT-RATE, validation split | 100 non-contrast chest CT volumes, 18 questions each (1,800 answers) | Yes/no per finding | Report-derived labels, as above |
| MR-ART motion | Image perception (volume) | MR-ART, OpenNeuro ds004173 | 436 T1-weighted volumes | Motion artifact: none or minimal / moderate / severe (ordinal) | Expert quality score 1 / 2 / 3 |
| BraTS sequence | Image perception (volume) | BraTS 2021 training set | 400 volumes (100 patients × 4 sequences) | T1 / T1c / T2 / FLAIR | Sequence given by the file name |

**CT-RATE.** CT-RATE pairs non-contrast chest CT volumes with their reports and 18 abnormality labels extracted from the reports by a text classifier. One reconstruction per scan was kept from the validation split. For the report task, 1,000 reports were sampled from all validation scans, and the model saw the Findings and Impressions sections. For the volume task, 100 scans were sampled from those whose volumes had been downloaded, so the two case sets overlap only in part. Each case was asked one yes/no question per finding.

**PadChest.** PadChest reports are in Spanish and are labelled with the findings they contain. Only physician-labelled reports were used, one per study, and 1,000 were sampled. Each report was asked one yes/no question for each of the 20 most frequent findings in that pool. A finding the report negates (for example "no pleural effusion" except in Spanish) counts as absent, so some questions test negation. Only the report text was used, not the images.

**GUDID.** Device records with an MRI safety status of Safe, Conditional or Unsafe were reservoir-sampled to 333 per class. The model saw the brand name, version or model number, company name, free-text description and up to two GMDN terms with their definitions. The status field itself was withheld but phrases in the description that state the status were not masked and occasionally included MR compatibility information. No specific access to device compatibility documents was provided, however. Therefore the task was to identify the device from its text and infer its MR safety from what the model already knows. For example, a ferromagnetic implant or a pacemaker leans toward MR Unsafe or Conditional, and a plastic or titanium item leans toward MR Safe. No guideline text was provided and nothing was retrieved. That makes it a knowledge-and-inference task, not a lookup.

**RANZCR CLiP.** One radiograph per patient was kept, and 500 were sampled at random. Each image was asked 11 yes/no questions covering endotracheal tube (normal, borderline, abnormal), nasogastric tube (normal, borderline, abnormal, incompletely imaged), central venous catheter (normal, borderline, abnormal) and Swan-Ganz catheter presence. Only the training labels are public, so these images may have appeared in model pre-training data.

**MR-ART.** All 436 T1-weighted volumes with an expert score were used: 129 none or minimal, 109 moderate and 198 severe.

**BraTS 2021.** One hundred patients with all four sequences were sampled, and each sequence became a separate case. The volumes are skull-stripped and co-registered.

| Model | Type | Checkpoint | Weights | Server | Max context (tokens) |
|---|---|---|---|---|---|
| Clef 27B | System 1 decision model, fine-tuned from Qwen3.8-27B | Cloudflare/clef via clef-vllm-spark (text); ollama.com/library/clef:27b (images) | BF16 (text); GGUF Q4_K_M, vision projector BF16 (images) | vLLM 0.28.0 (text); Ollama (images) | 16,384 |
| d1 3B | System 1 decision model (Liquid AI) | ollama.com/library/d1:3b | As released in the Ollama library | Ollama | 16,384 |
| Qwen3.8-27B | Generative VLM (Clef's backbone) | Qwen/Qwen3.8-27B-FP8 | FP8 | vLLM 0.32.0 | 32,768 |
| Gemma 4 31B | Generative VLM (Google) | google/gemma-4-31B-it | BF16 | vLLM | 32,768 |
| MedGemma 1.5 4B | Generative VLM, tuned on medical | google/medgemma-1.5-4b-it | BF16 | vLLM | 32,768 |

text and
images
(Google)

All models are multimodal and were used without fine-tuning by the authors. They were run one at a time on one NVIDIA DGX Spark (GB10 Grace Blackwell, 128 GB unified memory). Qwen3.8-27B was served with vLLM 0.32.0 with prefix caching on, a 32,768-token context, GPU memory utilization 0.7 and up to four images per prompt. Qwen3.8-27B reasons before answering by default, so thinking was turned off through the chat template (enable_thinking: false) and the first generated token was the answer. Gemma 4 31B (google/gemma-4-31B-it) and MedGemma 1.5 4B (google/medgemma-1.5-4b-it) were executed with the same code, cases, seed and prompts. Both were served at BF16 with vLLM, with the same context length, memory setting and image limit, and with thinking turned off where the chat template supports it.

**Clef and d1.** Clef's vLLM build accepts text only, so Clef was run two ways. For the three text tasks it was served at BF16 with vLLM 0.28.0 (clef-vllm-spark). For the four image tasks it was served with Ollama using its released Q4_K_M build. d1 was served with Ollama for all seven tasks. Both models were queried through the System 1 decision endpoint (/v1/systemone) with the harness's HTTP decision backend. Each request carried the same input text and images as for the generative model, plus every question of the case with its typed options and their one-line definitions. Clef and d1 score all options of all questions jointly in a single non-autoregressive pass and return a probability for each. These were renormalised to sum to one per question. They therefore needed one request per input, where the generative model needed one request per question.

**Prompt and readout.** For the three generative models, every question was sent as its own chat request. The system prompt was: "You are a medical decision model. Read the input and answer the question with the single letter of the best option. Output only the letter." The user message held the images, a short description of the input, the question and the lettered options (each option with a one-line definition where the task supplies one), ending with "Answer with one letter." For example, on the BraTS image type question, the user message ended:

```
Which MRI sequence is this?
A. T1
B. T1c
C. T2
D. FLAIR
Answer with one letter.
```

The visible output was a single token, for example "C". With log-probabilities requested, the server also returned the 20 most likely tokens at that position, for example (values illustrative):

```
"C" -0.22   "D" -1.70   " C" -4.10   "A" -5.30   "The" -6.00   …
```

Tokens were stripped of spaces and punctuation and upper-cased, so " c" counted as "C", and the highest log-probability was selected. Tokens that were not option letters, such as "The", were dropped, and a softmax over the four letters gave T1 0.01, T1c 0.00, T2 0.81 and FLAIR 0.18. The answer was T2, and its probability, 0.81, was the confidence used for calibration and selective accuracy. Yes/no questions were posed the same way: A. yes and B. no. Letters were used rather than class names because only the first token is scored, and class names can share a first token (MR Safe, MR Conditional and MR Unsafe all begin with "MR") or span several tokens. Clef and d1 instead return one probability per declared option, for example {"T1": 0.02, "T1c": 0.03, "T2": 0.78, "FLAIR": 0.17}, for all questions of a case in one response. Gemma 4 can write a short template marker before its answer even with thinking off, so for Gemma 4 and MedGemma up to 16 tokens were generated and the answer was read at the first generated token that was an option letter, from that position's top log-probabilities. This applied to 16 Gemma 4 answers and 1,236 MedGemma answers. No option letter was among the top 20 tokens for 354 MedGemma answers and 7 Gemma 4 answers (under 1% of answers); these were given equal probability across the options possible for reporting results.

## Image preprocessing and volume handling

**Radiographs.** RANZCR images were downsampled (Lanczos) so the longer side was at most 1,024 pixels and sent as one PNG per request.

**Volumes.** MR-ART, BraTS and CT-RATE volumes were read from NIfTI and given as axial slices in radiological orientation (anterior up, patient right on image left). Slices were selected every 5 mm between 10% and 90% of the volume's extent along z, up to 48 slices. MR intensities were clipped and scaled per volume to the 0.5th–99.5th percentiles of non-zero voxels. CT volumes stored as raw values were converted to Hounsfield units with the rescale slope and intercept from the CT-RATE metadata. Each CT slice was rendered in a lung window (center −600 HU, width 1,500 HU) and a mediastinal window (center 40 HU, width 400 HU), and both images went in one request. Each slice was resized to at most 512 or less (if native was less) pixels on its longer side. The text told the model that the input showed one axial slice, and for CT which windows were shown.

**Slice-level requests and pooling.** Following the paper's "slice" protocol, each slice was a separate request. This came to a mean of 42.0 requests per MR-ART volume (18,312 in total), 25.0 per BraTS volume (10,000 in total) and 47.7 per CT-RATE volume (4,772 in total). Per-slice probabilities were pooled into one answer per volume: the mean over slices for multi-class and ordinal questions (MR-ART and BraTS), and the maximum positive-class probability over slices for yes/no questions (CT-RATE findings).

$$p_{\text{volume}}(o) = \frac{1}{S}\sum_{s=1}^{S} p_s(o) \quad \text{(multi-class and ordinal)}$$

$$p_{\text{volume}}(\text{yes}) = \max_{s=1,\ldots,S} \; p_s(\text{yes}) \quad \text{(yes/no)}$$

### Metrics and statistical analysis

Metric definitions follow the reference paper.

- **AUROC (primary).** For yes/no questions, the AUROC of the positive-class probability. For multi-class and ordinal questions, the macro mean of one-vs-rest AUROCs over the classes present. Task AUROC is the mean over its questions, so RANZCR, CT-RATE and PadChest are the means of 11, 18 and 20 per-label AUROCs.
- **Chance-corrected balanced accuracy (BAcc).** Balanced accuracy of the highest-probability option, rescaled so that 0 is chance and 1 is perfect, then averaged over questions.
- **Expected calibration error (ECE).** Top-label ECE over 10 equal-width confidence bins, pooled over all answers in the task.
- **Selective accuracy.** Accuracy of all answers (100% coverage) and of the 50% of answers with the highest top-option probability.
- **Efficiency.** Median wall-clock latency per request, and GPU time per request from the cumulative NVML time (task total divided by requests). Latency and energy were measured in a separate run of the first 50 cases of each task, with no other workload on the device.

$$\text{BAcc}_{cc} = \frac{\text{BAcc} - 1/k}{1 - 1/k}$$

Here k is the number of classes present. A request that failed would have been scored as wrong, with the class prevalence as its probabilities. No request failed.

**Confidence intervals and model comparison.** The 95% CIs for task AUROC are percentile intervals from 1,000 bootstrap resamples of cases, with every question of a case resampled together. Models were compared with a paired bootstrap: 2,000 case resamples were drawn, the same resample was applied to both models, and ΔAUROC was recomputed each time.

Three comparisons were made on every task for a subset of models: Clef minus Qwen3.8-27B, d1 minus Qwen3.8-27B and d1 minus Clef. Figure 1 reports the percentile 95% CI of each difference, and a difference was called significant when its 95% CI excluded zero. No correction for multiple comparisons was applied.

# Results

## Discrimination, decision quality and calibration

All five models answered all 4,435 cases of the 7 tests with no failed request. Report retrieval was near perfect on CT-RATE and PadChest reports where the Qwen and Clef models reached AUROC 0.988–0.992 and d1 was 0.951–0.957 (Table 1). On the image and device tasks the models separated. Among the three original models, Clef had the highest AUROC on RANZCR, CT-RATE volumes and BraTS, and Qwen3.8-27B on GUDID and MR-ART, and d1 had the lowest AUROC on six of the seven tasks. On MR-ART it was close to Qwen3.8-27B and above Clef (Figure 1). Of the two added generative models, Gemma 4 31B had the highest AUROC of all five on GUDID (0.854) and MedGemma 1.5 4B on MR-ART (0.821).

| Task | Model | AUROC (95% CI) | BAcc | ECE | Acc@100% | Acc@50% |
|---|---|---|---|---|---|---|
| CT-RATE reports (n = 1,000, 18 labels) | Qwen3.8-27B | **0.992 (0.990–0.993)** | **0.903** | **0.015** | 0.954 | 0.993 |
| | Clef | 0.989 (0.988–0.991) | **0.903** | 0.026 | **0.963** | **0.999** |
| | d1 3B | 0.951 (0.947–0.954) | 0.838 | 0.036 | 0.918 | 0.990 |
| | Gemma 4 31B | 0.942 (0.938–0.947) | 0.899 | 0.045 | 0.955 | 0.946 |
| | MedGemma 1.5 4B | 0.934 (0.928–0.939) | 0.767 | 0.065 | 0.889 | 0.942 |
| PadChest reports (n = 1,000, 20 labels) | Qwen3.8-27B | **0.989 (0.985–0.992)** | **0.883** | **0.015** | 0.948 | 0.990 |
| | Clef | 0.988 (0.984–0.991) | 0.878 | 0.024 | **0.963** | **1.000** |
| | d1 3B | 0.957 (0.950–0.963) | 0.777 | 0.023 | 0.892 | 0.979 |
| | Gemma 4 31B | 0.933 (0.922–0.943) | 0.875 | 0.065 | 0.934 | 0.927 |
| | MedGemma 1.5 4B | 0.946 (0.937–0.955) | 0.681 | 0.107 | 0.756 | 0.882 |
| GUDID MR safety (n = 999) | Qwen3.8-27B | 0.762 (0.739–0.784) | 0.327 | 0.090 | 0.552 | 0.550 |
| | Clef | 0.741 (0.717–0.763) | 0.348 | 0.085 | 0.566 | 0.650 |
| | d1 3B | 0.595 (0.569–0.622) | 0.096 | **0.043** | 0.397 | 0.460 |
| | Gemma 4 31B | **0.854 (0.835–0.873)** | **0.518** | 0.285 | **0.679** | **0.848** |
| | MedGemma 1.5 4B | 0.615 (0.591–0.639) | 0.146 | 0.319 | 0.430 | 0.536 |
| RANZCR lines (n = 500, 11 labels) | Qwen3.8-27B | 0.766 (0.732–0.795) | **0.322** | 0.255 | 0.455 | 0.499 |
| | Clef | **0.812 (0.787–0.831)** | 0.099 | **0.025** | **0.871** | **0.970** |
| | d1 3B | 0.534 (0.491–0.582) | 0.035 | 0.171 | 0.451 | 0.412 |
| | Gemma 4 31B | 0.709 (0.663–0.749) | 0.204 | 0.281 | 0.698 | 0.845 |
| | MedGemma 1.5 4B | 0.689 (0.645–0.719) | 0.265 | 0.360 | 0.534 | 0.516 |
| CT-RATE 3D (n = 100, 18 labels) | Qwen3.8-27B | 0.668 (0.623–0.706) | 0.091 | 0.436 | 0.277 | 0.258 |
| | Clef | **0.749 (0.715–0.786)** | 0.198 | **0.072** | **0.778** | **0.872** |
| | d1 3B | 0.615 (0.564–0.667) | 0.074 | 0.095 | 0.535 | 0.582 |
| | Gemma 4 31B | 0.732 (0.692–0.767) | **0.210** | 0.545 | 0.443 | 0.250 |
| | MedGemma 1.5 4B | 0.707 (0.669–0.742) | 0.131 | 0.394 | 0.517 | 0.602 |
| MR-ART (n = 436) | Qwen3.8-27B | 0.709 (0.686–0.732) | 0.000 | 0.534 | 0.296 | 0.560 |
| | Clef | 0.625 (0.594–0.656) | 0.008 | 0.515 | 0.303 | 0.472 |
| | d1 3B | 0.702 (0.678–0.727) | 0.000 | 0.421 | 0.296 | 0.505 |
| | Gemma 4 31B | 0.589 (0.553–0.622) | 0.051 | 0.577 | 0.342 | 0.422 |
| | MedGemma 1.5 4B | **0.821 (0.797–0.845)** | **0.097** | **0.192** | **0.511** | **0.789** |
| BraTS (n = 400) | Qwen3.8-27B | 0.989 (0.983–0.994) | 0.487 | 0.168 | 0.615 | 0.930 |
| | Clef | **0.992 (0.988–0.996)** | **0.850** | 0.349 | **0.888** | **0.990** |
| | d1 3B | 0.797 (0.773–0.822) | 0.023 | 0.202 | 0.268 | 0.390 |
| | Gemma 4 31B | 0.973 (0.964–0.981) | 0.520 | **0.081** | 0.640 | 0.850 |
| | MedGemma 1.5 4B | 0.495 (0.465–0.527) | 0.000 | 0.531 | 0.250 | 0.235 |

**Table 1.** Task-level performance. AUROC with 95% case-bootstrap CI (1,000 resamples). BAcc = chance-corrected balanced accuracy (0 = chance). ECE = top-label expected calibration error (lower is better). Acc@100% and Acc@50% = accuracy of all answers and of the most confident half. Bold marks the best of the five models for each metric.

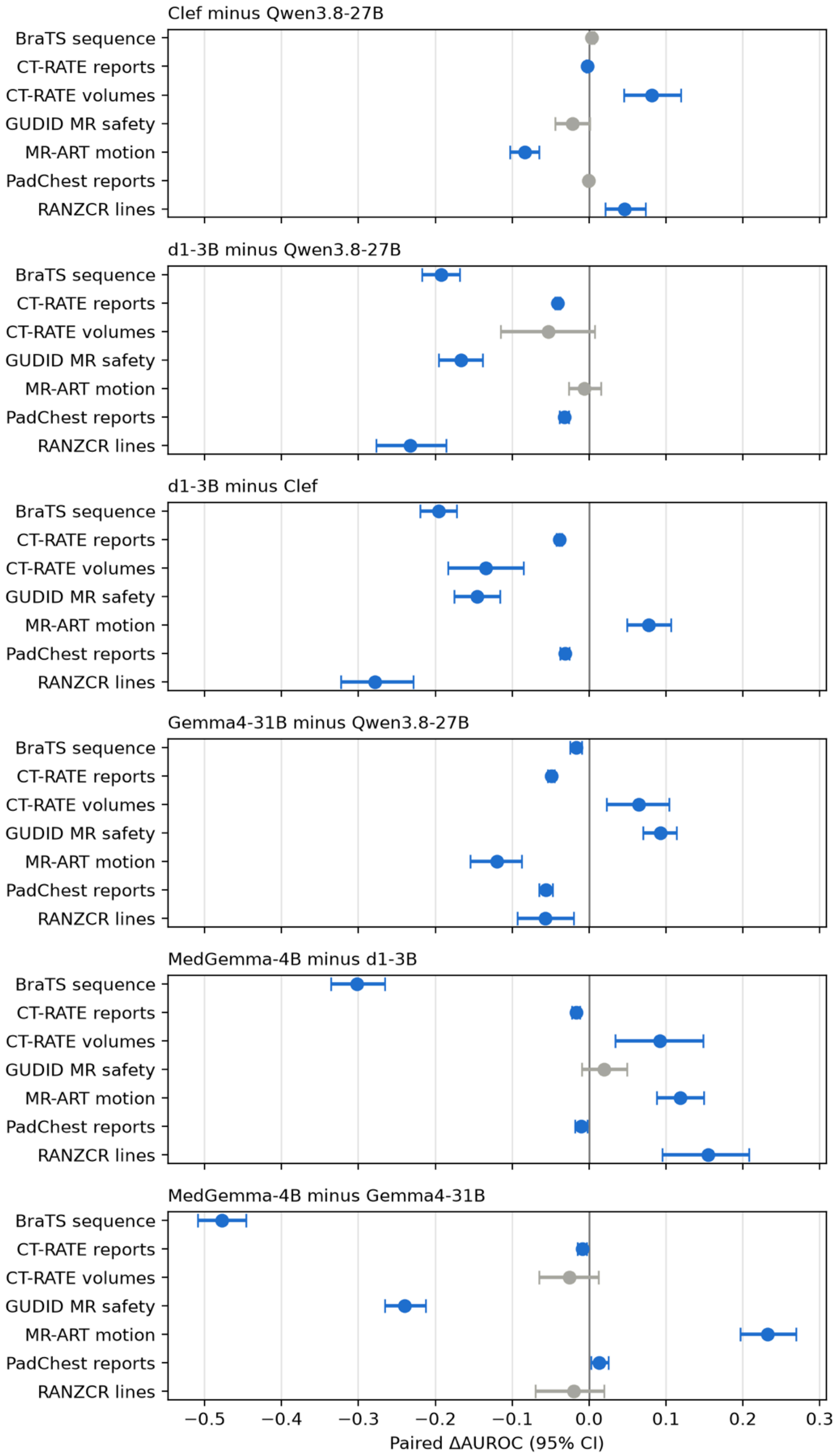

Figure 1. Paired ΔAUROC with 95% bootstrap CI (2,000 resamples) on each task, for, from top to bottom,

Clef minus Qwen3.8-27B, d1 minus Qwen3.8-27B, d1 minus Clef, Gemma 4 31B minus Qwen3.8-27B, MedGemma 1.5 4B minus d1 and MedGemma 1.5 4B minus Gemma 4 31B. Blue: the CI excludes zero, Gray: CI includes zero.

The AUROC gains did not always carry over to forced-choice accuracy. On CT-RATE volumes, Qwen3.8-27B ranked findings better than d1 (0.668 vs 0.615) but was less accurate (0.277 vs 0.535), because it answered "yes" to 88% of finding questions. On MR-ART, no model's top answer beat chance.

### Decision head versus backbone

Clef and Qwen3.8-27B start from the same weights, so their difference reflects the decision harness, together with the precision and serving stack used here. Their ranking did not differ on BraTS (paired ΔAUROC +0.003, 95% CI −0.000 to +0.007), GUDID (−0.021, −0.044 to +0.001) or PadChest reports (−0.001, −0.003 to +0.001). On CT-RATE reports Clef was lower by 0.002 (−0.004 to −0.001), which we view as too small a difference to matter at an AUROC of 0.99. Clef ranked higher on CT-RATE volumes (+0.082, +0.045 to +0.119) and RANZCR (+0.046, +0.021 to +0.074) and lower on MR-ART (−0.084, −0.103 to −0.065) (Figure 1).

The head changed decisions more than it changed ranking. On BraTS, Clef's top answer was right for 88.8% of volumes versus 61.5% for its backbone, and its chance-corrected balanced accuracy was 0.850 against 0.487. Its most confident half was 99.0% correct. Its calibration error was nevertheless higher (0.349 vs 0.168), because it was right more often than its probabilities implied. On GUDID, Clef was slightly more accurate (0.566 vs 0.552) with similar calibration (ECE 0.085 vs 0.090).

On RANZCR, Clef was the best-calibrated model in the study (ECE 0.025), and its most confident half was 97.0% correct. It reached that accuracy by answering "no" almost always: only 2.7% of its answers were "yes". On the five labels with fewer than 20 positives it never answered "yes". Its chance-corrected balanced accuracy over all labels was 0.099, below its backbone's 0.322. This is the opposite bias to its backbone, which answered "yes" to 65% of label questions (counting exact yes/no ties as "yes", as the accuracy metric does) against a mean label prevalence of 13%.

On MR-ART the head did not matter. Clef labelled 99.3% of scans "none or minimal", the same one-class collapse seen in Qwen3.8-27B and d1.

### CT findings

CT volumes showed the same pattern as RANZCR. Qwen3.8-27B answered "yes" to 88% of the 18 finding questions against a mean prevalence of 17%, so its accuracy was 0.277 and its ECE 0.436. Clef answered "yes" to 19%, with an accuracy of 0.778 and an ECE of 0.072, and it also ranked findings better. Its highest per-finding AUROCs were for pericardial effusion (0.971, 5 positives), pleural effusion (0.958, 9 positives) and cardiomegaly (0.932, 10 positives). Lung nodule, the most common finding (51 positives), was below chance for Clef and Qwen3.8-27B (0.43–0.44) and at chance for d1 (0.50).

### A 3B System 1 model

d1, with about one ninth of Clef's parameters, was close to the 27B models only on report text. Its AUROC was 0.951 on CT-RATE reports and 0.957 on PadChest reports, 0.03–0.04 below Clef, with chance-corrected balanced accuracy of 0.838 and 0.777. On the image and device tasks it fell further behind Clef: by 0.195 on BraTS, 0.145 on GUDID, 0.135 on CT-RATE volumes and 0.278 on RANZCR, where it was near chance (0.534). The exception was MR-ART, where it ranked higher than Clef (+0.078, +0.049 to +0.107) but still called every scan "none or minimal". On CT-RATE reports, its AUROC for medical material was 0.364, below chance, where the 27B models reached 0.955–0.993.

### Two additional generative models

Gemma 4 31B and MedGemma 1.5 4B were read out in the same way as Qwen3.8-27B (Table 1). Neither matched Qwen3.8-27B on report text. Gemma 4 reached AUROC 0.942 on CT-RATE reports and 0.933 on PadChest, and MedGemma 0.934 and 0.946, slightly below d1 (MedGemma minus d1 on CT-RATE reports, Δ −0.017, 95% CI −0.022 to −0.012). Gemma 4 ranked lower than Qwen3.8-27B on five of the seven tasks but higher on CT-RATE volumes (0.732 vs 0.668; Δ +0.064, +0.023 to +0.104) and GUDID, where its AUROC of 0.854 (Δ +0.093, +0.071 to +0.114) was the best of all five models and its answers were spread evenly across the three classes (recall 0.67–0.69, Table 4).

MedGemma, the smallest generative model and the only one tuned on medical data, ranked motion severity best of all five models (AUROC 0.821, against 0.589–0.709 for the others; Δ +0.118 over d1, +0.088 to +0.149). It was also above d1 on RANZCR (0.689 vs 0.534) and CT-RATE volumes (0.707 vs 0.615). Its decisions, however, collapsed onto single classes. It called 92.7% of MR-ART scans "severe", the opposite of the "none or minimal" collapse of the other models, never answered "MR Unsafe" on GUDID, and on BraTS answered T1c for 99.8% of volumes and ranked the sequences at chance (AUROC 0.495). Gemma 4 showed the reverse of Qwen3.8-27B's BraTS behaviour: it identified every FLAIR volume and gave FLAIR 56.5% of its answers (Table 4). On RANZCR, Gemma 4 answered "yes" to 31% of label questions and MedGemma to 57%.

### Throughput and compute time

Throughput was measured in a separate run of the first 50 cases of each task, with one model at a time and no other workload on the device (Table 2). On the three single-question tasks, Qwen was faster than Clef per request (median 0.13–0.21 s against 0.31–0.40 s) and used 1.8 to 2.5 less time. We note that for model availability reasons, we did have to use different serving models (Ollama vs vllm) which might account for this difference. On tasks with many questions per input the order reversed, because Clef answered all questions in one pass while the backbone made one call per question. Clef was 2.9 times faster than its backbone on CT-RATE reports (1.48 vs 4.24 s), 2.3 times faster on PadChest reports and 1.5 times faster per CT-RATE slice, with 1.5 to 2.7 times less energy per request. On RANZCR, with 11 questions, latency was within 10% and Clef used 16% more time. d1 was the fastest model on every task, with median latencies of 0.05–2.7 s and 3 to 9 times less time per request than Clef. One CT-RATE volume took 4.3 minutes with Qwen3.8-27B, 3.1 with Clef and 1.5 with d1. We report energy though that is close to proportional to time (there is some fixed baseline energy required).

| **Task** | **Model** | **Requests** | **Median latency (s)** | **Energy per request (J)** |
|---|---|---|---|---|
| CT-RATE reports | Qwen3.8-27B | 50 | 4.241 | 374.6 |
| | Clef | 50 | 1.480 | 137.2 |
| | d1 3B | 50 | 0.316 | 19.5 |
| PadChest reports | Qwen3.8-27B | 50 | 3.203 | 190.3 |
| | Clef | 50 | 1.399 | 127.9 |
| | d1 3B | 50 | 0.247 | 14.4 |
| GUDID MR safety | Qwen3.8-27B | 50 | 0.206 | 17.0 |
| | Clef | 50 | 0.404 | 30.4 |
| | d1 3B | 50 | 0.048 | 3.9 |
| RANZCR lines | Qwen3.8-27B | 50 | 2.627 | 186.2 |
| | Clef | 50 | 2.383 | 216.8 |
| | d1 3B | 50 | 1.020 | 54.9 |
| CT-RATE volumes | Qwen3.8-27B | 2,387 | 5.155 | 493.0 |
| | Clef | 2,387 | 3.406 | 339.6 |
| | d1 3B | 2,387 | 2.692 | 121.8 |
| MR-ART motion | Qwen3.8-27B | 2,100 | 0.128 | 10.9 |
| | Clef | 2,100 | 0.311 | 27.1 |
| | d1 3B | 2,100 | 0.078 | 4.4 |
| BraTS sequence | Qwen3.8-27B | 950 | 0.145 | 12.6 |
| | Clef | 950 | 0.363 | 32.1 |
| | d1 3B | 950 | 0.073 | 4.3 |

**Table 2.** Inference cost per request, from a separate run of the first 50 cases of each task. A request is one input sent once: one report, one device record, one radiograph or one slice. For the generative model a request with several questions covers one model call per question; for Clef and d1 it is one call answering all of them. Energy is the NVML counter for the whole GPU, so it includes idle and background draw. Gemma 4 31B and MedGemma 1.5 4B were not included in the timing run.

## Per-class behaviour and calibration

Qwen3.8-27B's forced-choice answers collapsed onto one or two classes, even where AUROC was high. Clef avoided the collapse on BraTS and GUDID but not on MR-ART, and d1 collapsed on all three tasks (Table 3). Table 4 gives the same breakdown for Gemma 4 31B and MedGemma 1.5 4B.

| Task | Class (n) | Recall, Qwen3.8-27B | Recall, Clef | Recall, d1 | Share, Qwen3.8-27B | Share, Clef | Share, d1 |
|---|---|---|---|---|---|---|---|
| BraTS sequence | T1 (100) | 1.00 | 0.99 | 0.00 | 42.2% | 34.0% | 0.0% |
| | T1c (100) | 0.46 | 0.66 | 1.00 | 11.5% | 17.0% | 97.8% |
| | T2 (100) | 1.00 | 1.00 | 0.00 | 46.2% | 26.5% | 0.0% |
| | FLAIR (100) | 0.00 | 0.90 | 0.07 | 0.0% | 22.5% | 2.2% |
| GUDID MR safety | MR Safe (333) | 0.41 | 0.57 | 0.46 | 14.2% | 31.2% | 32.1% |
| | MR Conditional (333) | 0.41 | 0.61 | 0.05 | 28.5% | 41.5% | 4.3% |
| | MR Unsafe (333) | 0.83 | 0.51 | 0.68 | 57.3% | 27.2% | 63.6% |
| MR-ART motion | None or minimal (129) | 1.00 | 1.00 | 1.00 | 100.0% | 99.3% | 100.0 % |
| | Moderate (109) | 0.00 | 0.00 | 0.00 | 0.0% | 0.0% | 0.0% |
| | Severe (198) | 0.00 | 0.01 | 0.00 | 0.0% | 0.7% | 0.0% |

**Table 3.** Recall per class and share of all answers given to each class, for the three multi-class tasks. Classes are balanced in GUDID and BraTS.

| Task | Class (n) | Recall, Gemma 4 31B | Recall, MedGemma 1.5 4B | Share, Gemma 4 31B | Share, MedGemma 1.5 4B |
|---|---|---|---|---|---|
| BraTS sequence | T1 (100) | 0.59 | 0.00 | 19.2% | 0.2% |
| | T1c (100) | 0.14 | 1.00 | 3.5% | 99.8% |
| | T2 (100) | 0.83 | 0.00 | 20.8% | 0.0% |
| | FLAIR (100) | 1.00 | 0.00 | 56.5% | 0.0% |
| GUDID MR safety | MR Safe (333) | 0.67 | 0.35 | 27.3% | 22.3% |
| | MR Conditional (333) | 0.68 | 0.94 | 37.8% | 77.7% |
| | MR Unsafe (333) | 0.69 | 0.00 | 34.8% | 0.0% |
| MR-ART motion | None or minimal (129) | 1.00 | 0.19 | 95.4% | 7.3% |
| | Moderate (109) | 0.00 | 0.00 | 0.0% | 0.0% |
| | Severe (198) | 0.10 | 1.00 | 4.6% | 92.7% |

**Table 4.** Recall per class and share of all answers given to each class for Gemma 4 31B and MedGemma 1.5 4B, for the three multi-class tasks (same layout as Table 3).

**BraTS.** Qwen3.8-27B never chose FLAIR. It labelled 84 of 100 FLAIR volumes as T2 (the other 16 T1) and 53 of 100 T1c volumes T1. Its near-perfect AUROC (0.989) shows that the per-class probabilities still ranked the sequences correctly. The errors come from the argmax decision, not from missing information. Clef, with the same ranking (0.992), chose every sequence. It identified 90 of 100 FLAIR volumes and 99 or more of 100 T1 and T2 volumes. Its main error was calling 33 of 100 T1c volumes T1, so it was unsure about contrast enhancement on single slices. d1 answered T1c for 391 of the 400 volumes.

**GUDID.** Qwen3.8-27B favoured "MR Unsafe" (57% of answers) and was less confident than it was accurate (mean top-option probability 0.47 against 55% accuracy). Clef spread its answers more evenly (31%, 42% and 27%), with recall of 0.51 to 0.61 for each class. d1 almost never answered "MR Conditional" (4% of answers) and called 245 of 333 conditional devices "MR Unsafe". Its ECE was the lowest in the study (0.043) only because its confidence was as low as its accuracy (mean top-option probability 0.43 against 40% accuracy).

**MR-ART.** Clef, its backbone and d1 called almost every scan "none or minimal" with a mean top-option probability of 0.72–0.83, although 70% of scans had moderate or severe motion. Mean pooling over about 42 slices per volume may dilute artifact visible on only some slices. The AUROCs of 0.63–0.71 show that the probability of "none or minimal" still fell with increasing motion, so a tuned threshold could recover some decision value. Gemma 4 31B did the same (95.4% "none or minimal"), whereas MedGemma 1.5 4B called 92.7% of scans "severe" while ranking motion best of all five models (AUROC 0.821).

**RANZCR.** Both 27B models ranked endotracheal and nasogastric tube labels well (AUROC 0.75–0.97) but central venous catheter position near chance (0.50–0.56, Figure 2). The CVC labels are the most common ones. Clef ranked 9 of the 11 labels higher than its backbone, including all three ETT labels, NGT incompletely imaged (0.949 vs 0.820, 23 positives) and Swan-Ganz present (0.899 vs 0.762, 8 positives). d1 was near chance on every label (0.48–0.58). Qwen3.8-27B answered "yes" to 65% of label questions and d1 to 55%, against a mean label prevalence of 13%. Gemma 4 31B and MedGemma 1.5 4B ranked ETT labels at 0.73–0.95 and 0.81–0.96 and NGT labels at 0.49–0.77 and 0.63–0.71, were near chance on CVC labels (0.52–0.56) like every other model, and answered "yes" to 31% and 57% of label questions.

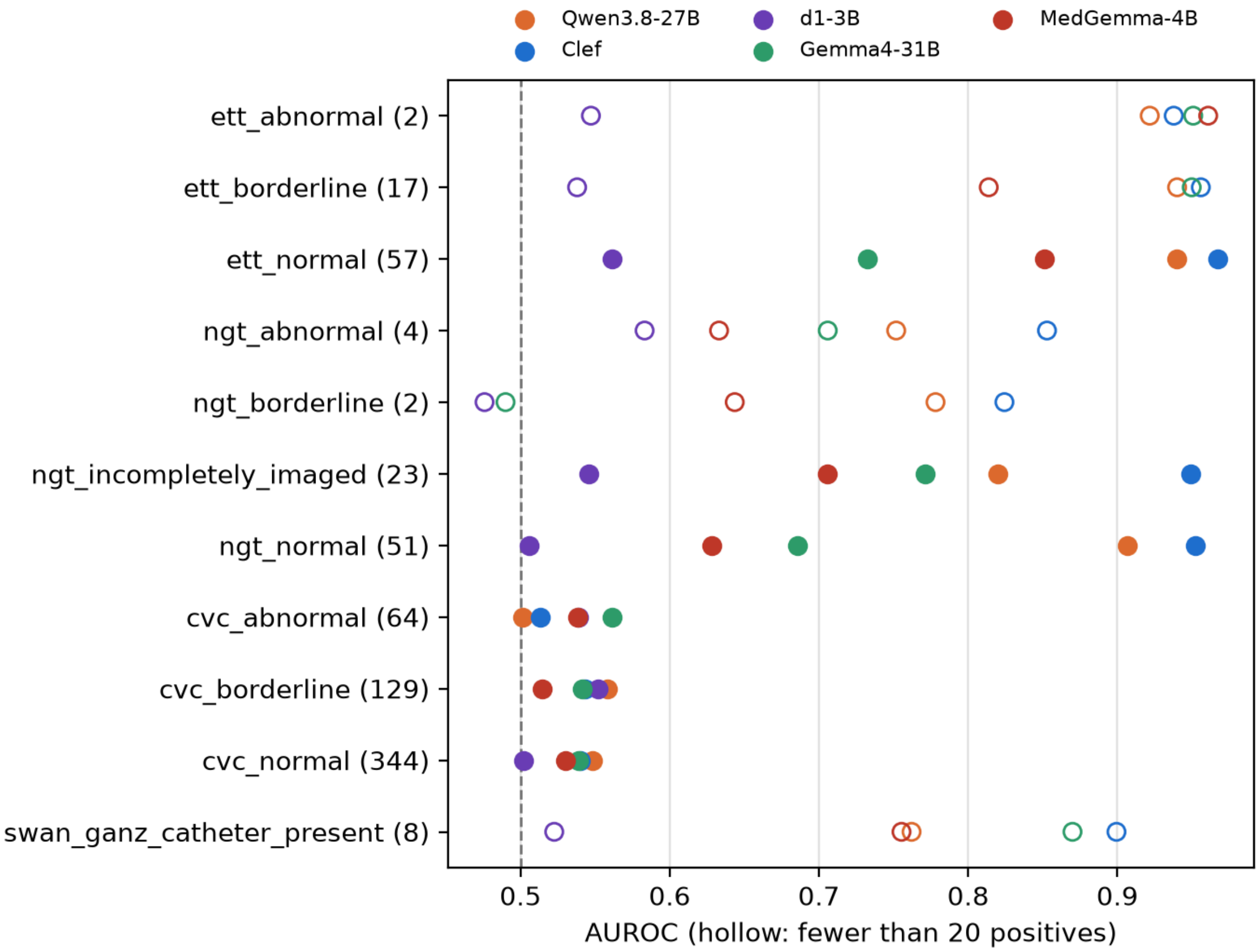

Figure 2. Per-label AUROC on RANZCR CLiP for all five models. Numbers in parentheses are positives among 500 radiographs; hollow markers have fewer than 20 positives. The dashed line is chance.

## Discussion

Three original models ranked report text and MRI sequences almost perfectly. What the head changed was the decisions: on MRI sequences, chest radiograph lines and CT findings, Clef's top answers and confidence were usable where its Qwen3.8's were not. Model size mattered more than the head for ranking. The 3B d1 was close to the 27B models on text but fell behind on every image and device task except motion where no model performed well.

### Ranking is not deciding

The main finding is the gap between discrimination and decisions. A System 1 model is useful only if its top answer is confident so that it can be automated, else referred for human intervention. The LLM readout failed that test on most image tasks. On BraTS, Qwen3.8-27B ranked the four sequences almost perfectly (AUROC 0.989) but never answered FLAIR: its FLAIR probability was highest on FLAIR volumes, but its T2 probability was higher still, so the top choice was T2. Ranking depends on each class's probability across cases, while the top answer depends on how the classes compare within a case. On MR-ART, the probability of "none or minimal" fell with motion severity, yet every model answered "none or minimal" for nearly every scan. On RANZCR and CT-RATE volumes, Qwen3.8-27B answered "yes" to 65% and 88% of questions against prevalence of 13% and 17%. In each case the information was in the probabilities, but the decision rule ignored it.

This pattern points to a biased option prior rather than missing knowledge, and the Clef results support that reading. A first-token letter readout inherits whatever preference the model has for a letter, a class name or "yes". A head trained to score options jointly removes that preference. With the same ranking on BraTS, Clef's chance-corrected balanced accuracy was 0.850 against 0.487 for its backbone. Yang, Vrettos et al. reported the same effect from the other side: five of 16 significant advantages of a System 1 head over its backbone disappeared or reversed once the backbone readout was corrected for its prior option. The report tasks showed no such bias, because the answer was stated in the text: Qwen3.8-27B answered "yes" to 22% of CT-RATE report questions against a prevalence of 20%.

The head did not remove every bias. On RANZCR, Clef traded its backbone's "yes" bias for a "no" bias. It was well calibrated (ECE 0.025) but never answered "yes" to any of the five rarest labels. This might be viewed as the model having learned to answer "no" but since there was no fine-tuning to tasks, we think this less likely. A calibrated probability with the default decision rule still misses rare findings, so per-label thresholds are needed whatever the readout. On MR-ART, no readout made usable decisions, and Clef also ranked worse than its backbone. That points to perception or pooling rather than the readout: averaging about 42 slices, most without visible motion, may wash out the artifact.

### A generative model constrained to one token

Reading a generative model output as a single token is not how LLMs are normally used. In the usual use, the LLM generates a free-text answer consisting of many token/words, and any confidence is output as words. Further, each word in the answer requires a forward pass through the model and the answer text is autoregressive with possible extra iterations if reasoning is enabled which significantly increases compute energy consumption and cost. Here the answer was forced into one token/letter, and the confidence was taken from the token probabilities that the server computes anyway. This approach turns any open generative model into a decision model without training, at a cost per task close to that of a System 1 model, since both make one pass over the input. Read out this way, Qwen3.8-27B ranked within 0.03 AUROC of Clef on four of the seven tasks and higher on MR-ART.

The readout has limits that a decision model is built to remove. Options must be mapped to single tokens, so the answer depends on letters and inherits the model's preference for some letters or for "yes". This might account for the 65% and 88% "yes" rates on RANZCR and CT-RATE volumes and lack of FLAIR answers on BraTS or it might simply be that the model did not know about those options. Probability can fall on tokens that are not options and has to be normalized away. Each question needs its own call, so several questions about one input will cost more than a single pass. The token probabilities were never trained to be calibrated, and a model built to reason was used with reasoning turned off. The method also needs access to log-probabilities, which self-hosted servers provide but many hosted APIs restrict. A System 1 model scores the declared options directly, answers all questions of a case together and is trained for calibrated probabilities. Those properties are what separated Clef from its backbone here: decisions and calibration rather than ranking. A site that has only a generative model may recover much of that difference with a single-token readout and a short local calibration step, such as a per-option bias correction or per-label thresholds.

### Comparison with the reference study

In the reference study, Clef reached an AUROC of 0.71 on MR-ART and 0.58 on GUDID. Here it reached 0.625 and 0.741. The lower MR-ART value may come from the 4-bit Q4_K_M build used for the image tasks here, from slice sampling and intensity scaling, or from case selection, and we did not pursue the cause. Qwen3.8 reached 0.709, close to the reference study's Clef.

On GUDID, both 27B models exceeded the reference study's best AUROC of 0.66, and d1 did not (0.595). This should not be read as better MR safety judgement. The classes were balanced here, and the device descriptions were not masked, so some records stated the MR status outright. A run with mask_mr_terms enabled is needed before the GUDID result can be compared.

### A small System 1 model

d1 shows where a small model may be enough. On report text, where the answer is stated, a 3B model came within 0.04 AUROC of the 27B models. Where the answer had to be seen in an image or inferred from a device description, it fell behind by 0.13 to 0.28 and its decisions collapsed onto one class. For labelling findings from reports a small model may be adequate; for image interpretation these results do not support it.

### Other generative backbones

The two added models show that the single-token readout is not specific to Qwen3.8-27B. Both answered with an option letter in more than 99% of requests with no change to the prompt, and both ranked cases usefully on most tasks. They also show that no backbone is best everywhere. Gemma 4 31B was the best of all five models on MR safety and better than Qwen3.8-27B on CT volumes, but it was weaker than Qwen3.8-27B on report text, so a larger or newer model is not automatically better at retrieval. MedGemma 1.5 4B, about one eighth the size of Gemma 4, ranked motion severity better than any other model, consistent with its training on medical images, yet it ranked MRI sequences at chance. Each backbone also brought its own option bias. Qwen3.8-27B never chose FLAIR, Gemma 4 chose FLAIR for more than half of all volumes, and MedGemma chose T1c for almost every volume, "severe" for almost every MR-ART scan and never "MR Unsafe". Ranking quality and decision bias therefore have to be checked per model and per task. Because the readout costs about as much as a System 1 model and needs no training, several open models can be screened on local cases and the best one chosen for each task.

### Where the speed of a System 1 model comes from

The reference study found System 1 models 11 to 76 times faster than their backbones, but the backbones there reasoned before answering. Here the backbone had thinking turned off and generated one token. It was faster per request than Clef on every single-question task. Clef was faster only where one pass replaced many calls: 2.9 times on CT-RATE reports (18 questions), 2.3 times on PadChest reports (20 questions) and 1.5 times per CT slice (18 questions). Under these conditions the speed of a System 1 model comes from skipping the reasoning trace and from answering several questions in one pass, not from the head itself. Model size mattered more: d1 used 3 to 9 times less energy per request than Clef. The serving stacks differed, so the exact ratios are specific to this setup.

### Implications for System 1 use in radiology

Neither a general-purpose LLM nor a dedicated decision model, used zero-shot, is satisfactory for image-based tasks but may be useful for many text-based tasks. Clef gives more usable top answers on MRI sequences and better calibration on chest radiograph and CT findings, which makes it the better starting point. The 3B d1 model is a reasonable choice only for report text. All of them need a local calibration step for most tasks: a few hundred labelled cases per task to set per-class or per-label thresholds and check calibration before deployment. We demonstrate this can be performed on minimal local hardware. Because the single-token readout worked across three model families, several open models can be screened on local cases and the best one chosen for each task.

## Limitations

There are several limitations to our study including:

- **Model, precision and serving confounded.** Clef ran at BF16 under vLLM for the text tasks and at Q4_K_M under Ollama for the image tasks, and its backbone at FP8 under vLLM, so the head-versus-backbone comparison also includes a precision and serving difference, and latency and energy compare serving stacks as well as models. d1 differs from Clef in size, base model and training, so its results cannot be attributed to size alone. No request failed under any context limit.
- **Single readout and prompt.** Generative results reflect one zero-shot prompt and first-token letter probabilities with thinking off. No prompt variation, option-order permutation or reasoning mode was tested.
- **Second session.** Gemma 4 31B and MedGemma 1.5 4B were run about two weeks after the other models, with the same code, cases, seed and prompts but a newer vLLM image. For these two models the answer was read at the first option letter within the first 16 generated tokens rather than strictly at the first token.
- **Volume protocol.** Volumes were sent one slice at a time and pooled by the mean for MRI and by the maximum for CT findings. Other pooling rules (top-3 mean) and multi-slice or montage inputs may change the volume results, especially the MR-ART class collapse and the generative model's "yes" rate on CT.
- **Label and leakage caveats.** GUDID device descriptions were not masked and sometimes state the MR status outright. CT-RATE labels were extracted from the reports by a text classifier, so the report task partly measures agreement with that classifier, and the volume task is scored against report-derived labels. RANZCR uses the public training split, which may be in pre-training data. BraTS volumes are skull-stripped and co-registered, which removes some acquisition cues.

- **Small positive counts.** Five of the 11 RANZCR labels and 12 of the 18 CT-RATE volume findings have fewer than 20 positives, so their per-label AUROCs are imprecise. The CT-RATE volume task has only 100 cases.
- **Scope.** PadChest was evaluated from the report text only, not from the radiographs.
- **Statistics.** The 42 paired comparisons (six model pairs on seven tasks) are not corrected for multiplicity.
- **Energy measurement.** Energy is whole-GPU NVML readings on a unified-memory system and includes idle and background draw. It is suitable for comparing models on this host, not as an absolute cost per inference.
- **Timing.** Latency and energy come from a separate 50-case run. Accuracy and calibration of predictions come from the full run.

## Conclusion

A System 1 decision head added little discrimination over its own backbone on report text, MR safety and MRI sequences, gained on chest radiograph lines and CT findings, and lost on MRI motion. It produced more usable decisions, most clearly on MRI sequence identification, and much better calibration on chest radiograph and CT findings. A 3B System 1 model matched the 27B models only on report text. Two further generative models, Gemma 4 31B and MedGemma 1.5 4B, could be read out the same way without changes; each was the best of all five models on one task (MR safety and motion ranking, respectively) and weaker or strongly biased on others. No readout made usable decisions on motion severity, and rare findings were missed by a calibrated head as well as by a biased backbone. Probability readouts need local calibration and decision thresholds before they can act as System 1 components in radiology.